\documentclass[11pt]{article}
\usepackage[preprint]{acl}

\usepackage{times}
\usepackage{latexsym}
\usepackage[T1]{fontenc}
\usepackage[utf8]{inputenc}
\usepackage{microtype}
\usepackage{inconsolata}
\usepackage{amsmath,amssymb,amsfonts}
\usepackage{booktabs}
\usepackage{graphicx}
\usepackage{algorithm}
\usepackage{algorithmic}
\usepackage{multirow}
\usepackage{subcaption}
\usepackage{xcolor}
\usepackage{array}
\usepackage{tabularx}
\usepackage{makecell}
\usepackage{float}
\usepackage{tikz}
\usepackage{pgfplots}
\usepackage{placeins}
\usepackage{balance}
\pgfplotsset{compat=1.17}
\usetikzlibrary{arrows.meta,positioning,fit,calc}
\makeatletter
\@ifpackageloaded{hyperref}{\hypersetup{
    colorlinks=true,
    citecolor=darkblue,
    linkcolor=darkblue,
    urlcolor=darkblue
}}{}
\makeatother

\newcommand{\ours}{\textsc{ProxyMix}}

\newcommand{\cT}{\mathcal{T}}

\title{Dynamic Proxy-Mixing: Transferring Replay Controllers from Small to Large Models for Continual Instruction Tuning}

\author{
\textbf{Ibne Farabi Shihab}\textsuperscript{1}%
\thanks{Equal contribution.}%
\thanks{Corresponding author: \texttt{ishihab@iastate.edu}.}
\and
\textbf{Fariya Afrin}\textsuperscript{2}\footnotemark[1]
\and
\textbf{Anuj Sharma}\textsuperscript{3}
\\[2pt]
\textsuperscript{1}Department of Computer Science, Iowa State University \\
\textsuperscript{2}Department of Computer Science, Kalinga Institute of Industrial Technology \\
\textsuperscript{3}Department of Civil, Construction \& Environmental Engineering, Iowa State University \\
\texttt{ishihab@iastate.edu}
}

\begin{document}
\maketitle

\begin{abstract}
Continual instruction tuning updates a language model through a sequence of new domains, yet each update can progressively erode previously learned capabilities and alignment behavior. Replay is the standard mitigation, but fixed replay ratios are inherently limited because the optimal mixture varies with the current domain, the training stage, and the evolving vulnerability of prior behaviors. We propose PROXYMIX, a framework that learns a dynamic replay controller on a small proxy model and transfers the frozen controller to a larger target. The controller never observes future tasks and constructs its state from normalized validation losses and their temporal dynamics, producing a masked mixture over the current task and accessible replay buffers. Our core empirical hypothesis is \textit{forgetting mirroring}: task vulnerability rankings remain largely consistent across model scales even when absolute loss magnitudes differ. We validate this assumption empirically before transferring controllers across scales. On LLaMA-3-8B across five continual instruction tuning sequences, PROXYMIX improves average accuracy by 3.4 points, reduces final forgetting by 3.5 points, and raises safety score by 5.8 points over the strongest non-oracle baseline, at roughly 50x lower policy learning cost than Oracle Target RL. The framework is leakage free and architecture independent at the interface level, and we also identify settings where the proxy assumption breaks down, highlighting limitations for robust deployment.

\end{abstract}

\section{Introduction}

\noindent Language models are rarely adapted in a single step. A general-purpose chat model is typically fine-tuned sequentially across a stream of heterogeneous domains. It ranges from code generation to medical question answering, legal drafting, creative writing, multilingual instruction following, and evolving safety policies. While this staged adaptation is computationally efficient and avoids retraining from scratch, it introduces a persistent continual-learning challenge: each update improves performance on the current domain at the potential cost of degrading previously acquired capabilities and behavioral constraints.

\noindent In instruction-tuned systems, this degradation is not limited to superficial accuracy drops. What is forgotten may include refusal behaviors, conversational norms, multi-turn reasoning structures, or structured instruction-following patterns ~\citep{qi2023fine, zhan2024removing}. We refer to this broader phenomenon as \emph{alignment drift}, and it represents the central stability failure mode addressed in this work. Experience replay is the most widely used mitigation strategy ~\citep{chaudhry2019tiny, rebuffi2017icarl}, where data from previous tasks is interleaved with current-task training. However, fixed replay ratios are inherently limited: the optimal replay mixture varies across domains, training stages, and the evolving vulnerability of prior behaviors. A static allocation may under-protect safety-critical or highly fragile behaviors, while over-allocating resources in stable or low-interference transitions.

\noindent These limitations motivate the need for a dynamic replay controller that adapts the mixture over time. Ideally, such a controller would allocate higher replay to tasks exhibiting rising validation loss, reduce replay when prior tasks remain stable, and prioritize retention of safety-critical behaviors. However, learning such a controller directly on large target models is computationally prohibitive, as each reinforcement learning episode requires a full continual training trajectory. This motivates a proxy-to-target paradigm: a smaller proxy model can be used to efficiently explore replay policies, identifying vulnerability patterns and effective mixture strategies, which are then transferred to a larger target model.

\noindent We introduce \textsc{PROXYMIX}, a two-stage framework for dynamic replay control in continual instruction tuning. PROXYMIX introduces a proxy-to-target framework for learning dynamic replay policies on small proxy models and transferring them to large target models for continual instruction tuning, achieving near-oracle performance at substantially lower optimization cost than Oracle Target RL. In the proxy stage, a lightweight model is trained over a sequence of tasks while a policy learns to select mixtures over the current task and replay buffers. The learned controller operates under a strict information constraint: at stage $k$, it observes only normalized validation signals and loss trends for tasks $T_1 \dots T_k$, while future tasks remain fully masked in state, action, and reward. This prevents leakage that would otherwise reduce the problem to offline curriculum optimization rather than continual learning. In the target stage, the learned policy is frozen and directly applied to a larger model, with decisions driven by relative drift signals rather than absolute loss scales.

\noindent A key empirical observation underlying our approach is \emph{forgetting mirroring}: tasks that are most susceptible to forgetting in the proxy model tend to exhibit similar vulnerability in the target model under shared replay conditions. While not universal, this property is particularly strong under interference-heavy or style-shifting task transitions and provides a practical basis for transferring replay control across scales. When combined with limited validation on the target model, it enables effective and computationally efficient policy reuse.

Our contributions are as follows:
\begin{itemize}
    \item We propose \ours{}, a leakage-free proxy-to-target framework for dynamic replay control in continual instruction tuning.
    
    \item We formalize a masked state--action--reward formulation that enforces strict continual learning constraints during policy optimization.
    
    \item We introduce and empirically investigate the forgetting-mirroring hypothesis across model scales and task sequences.
    
    \item We show that proxy-learned policies consistently outperform fixed and heuristic replay baselines while substantially reducing optimization cost.
    
    \item We conduct ablation studies isolating the effects of normalization, temporal trend signals, policy expressivity, and safety-aware rewards, and we include a leakage-control analysis showing why future-task masking is required for a valid continual-learning evaluation..
\end{itemize}

\noindent Overall, \textsc{PROXYMIX} reframes replay not as a static hyperparameter but as a learned, transferable control policy that adapts to the evolving structure of alignment drift across continual instruction tuning.
\section{Related Work}

Continual learning has traditionally addressed the stability–plasticity tradeoff through regularization, architectural isolation, and replay-based approaches. Regularization methods such as EWC and SI discourage changes to parameters estimated to be important for earlier tasks~\citep{kirkpatrick2017overcoming,zenke2017continual}. Architecture-based methods reserve or expand capacity across tasks~\citep{rusu2016progressive}. Replay-based methods store or synthesize old examples and train on them alongside new data~\citep{lopez2017gradient,rebuffi2017icarl,chaudhry2019tiny}. Our method belongs to the replay family, but instead of treating replay allocation as a fixed sampling rule, it treats allocation as a sequential control problem that adapts to the model's current state.

Continual instruction tuning sharpens these classical concerns and introduces additional constraints. CITB formulates continual instruction tuning as a benchmarked setting with long instruction streams~\citep{zhang2023citb}, while TRACE evaluates continual learning for aligned LLMs across domain-specific, multilingual, code, and mathematical tasks and measures degradation beyond task accuracy~\citep{wang2023trace}. Recent surveys emphasize that LLM continual learning spans continual pre-training, continual fine-tuning, and continual alignment, with different forgetting modes at each stage~\citep{shi2025continual}. Work on aligned models is particularly relevant because it demonstrates that domain-specific fine-tuning can compromise safety behavior even when the new data are not explicitly adversarial~\citep{qi2023fine,zhan2024removing}. These results motivate evaluating retention not only as held-out task accuracy but also as preservation of instruction-following and refusal behavior, both of which we measure.Additional discussion of closely related replay and proxy-based methods is provided in Appendix~\ref{app:related}, while Appendix~\ref{app:positioning} summarizes how \textsc{PROXYMIX} compares to prior approaches across key methodological dimensions.

\section{Problem Setting}

We study a sequence of instruction-tuning tasks \(\cT_1,\ldots,\cT_T\) presented one after another. At stage \(k\), the learner receives current-task data from \(\cT_k\) and may replay data from \(\cT_1,\ldots,\cT_{k-1}\). It cannot observe validation examples, rewards, or replay buffers for future tasks. This restriction is essential for a faithful continual-learning evaluation: a method that peeks at \(\cT_{k+1}\) when choosing replay for \(\cT_k\) is no longer solving the continual problem but a curriculum problem disguised as one.

To compare models across scales, we need a measure of forgetting that does not depend on absolute scores. Let \(A_i^{(m)}(k)\) denote the score of model scale \(m\) on task \(\cT_i\) after completing stage \(k\). For \(j>i\), the relative forgetting from completing \(\cT_i\) to completing \(\cT_j\) is
\begin{equation}
 \phi_{i\to j}^{(m)}=
 \frac{A_i^{(m)}(i)-A_i^{(m)}(j)}{\max(A_i^{(m)}(i),\epsilon_A)},
 \label{eq:pairwise_forgetting}
\end{equation}
where \(\epsilon_A\) prevents division by zero. Positive values indicate degradation from the checkpoint immediately after learning \(\cT_i\). We use this pairwise quantity to compare forgetting rankings across scale, since absolute losses differ between proxy and target even when their relative profiles agree.

For final evaluation, we report average accuracy,
\[
\mathrm{AA}=T^{-1}\sum_i A_i(T),
\]
and final forgetting, defined as the average drop from each task's best previous checkpoint:
\begin{equation}
\begin{split}
\mathrm{FGT}=
\frac{1}{T-1}\sum_{i=1}^{T-1}
\max\!\bigl(
0,\,
\max_{r\in\{i,\ldots,T-1\}}
A_i(r)
\\
{}- A_i(T)
\bigr).
\end{split}
\end{equation}
We also report new-task accuracy,
\[
\mathrm{NTA}=T^{-1}\sum_i A_i(i),
\]
and backward transfer,
\[
\mathrm{BWT}=
(T-1)^{-1}\sum_{i=1}^{T-1}
\bigl(A_i(T)-A_i(i)\bigr),
\]
which together separate plasticity from stability. Safety score is reported as a balanced score that combines harmful-prompt refusal accuracy with a penalty for benign over-refusal, since harmful-only refusal accuracy can be gamed by blanket refusal; the full definition appears in Appendix~\textcolor{darkblue}{\ref{app:safety}}.

\section{Forgetting Mirroring}
\label{sec:mirroring}

Proxy transfer is useful only if the proxy reveals something stable about the target. Before introducing the controller, we therefore stop to ask a diagnostic question: when a proxy and target are trained on the same sequence under matched conditions, do they agree on which tasks are fragile? If they do, even approximately, then a controller trained on the proxy has a reasonable chance of allocating replay sensibly when transferred. If they do not, no policy learned on the proxy will transfer reliably, and the entire framework is misconceived.

To make the question concrete, we train the proxy and target with the same simple replay protocol on each sequence in Table~\ref{tab:task_sequences} and compute pairwise forgetting values \(\phi_{i\to j}\) for both models. We then compare the proxy and target rankings with Spearman correlation. The test asks whether the proxy identifies which tasks are vulnerable, not whether it matches the target's absolute scores. The sequences span sharp domain shifts (S1), longer streams with stylistic and topical variety (S2, S4), a safety-critical stream where a small early dataset must be preserved through subsequent updates (S3), and a closely related-domain stream where interference is mild (S5). Together they give the diagnostic a chance to fail in several different ways.

\begin{table}[t]
\centering
\small
\begin{tabular}{ll}
\toprule
\textbf{ID} & \textbf{Tasks in order} \\
\midrule
S1 & Code \(\to\) Math \(\to\) Chat \\
S2 & Chat \(\to\) Medical \(\to\) Legal \(\to\) Code \\
S3 & Safety \(\to\) Code \(\to\) Writing \(\to\) Math \\
S4 & Summarization \(\to\) QA \(\to\) Code \(\to\) Chat \(\to\) Math \\
S5 & Code \(\to\) Python \(\to\) Rust \(\to\) Math \\
\bottomrule
\end{tabular}
\caption{Continual instruction-tuning sequences used throughout the paper. They include both sharp domain shifts and related-domain transitions, and S3 contains an early safety dataset that must be preserved through subsequent updates.}
\label{tab:task_sequences}
\end{table}

The results in Table~\ref{tab:mirroring} show what the diagnostic finds. Across scale pairs, the Spearman correlation of forgetting rankings is consistently high, and the confidence intervals exclude weak or random agreement. Larger proxies generally track the target more closely, with the 410M--6.9B and 1.4B--8B pairings both reaching high correlations. Even the smallest proxy carries enough signal to be useful. Appendix~\ref{app:scatter} visualizes the same finding as a per-task rank scatter: points cluster near the diagonal, and the off-diagonal departures correspond to tasks where the two models genuinely disagree about vulnerability rather than to noise.

\begin{table}[t]
\centering
\small
\begin{tabular}{lccc}
\toprule
\textbf{Proxy \(\to\) target} & \textbf{mean \(\rho\) }& \textbf{95\% CI} & \textbf{exact \(p\) }\\
\midrule
160M \(\to\) 6.9B & 0.81 & [0.70, 0.89] & 0.006 \\
410M \(\to\) 6.9B & 0.90 & [0.82, 0.95] & 0.002 \\
1.4B \(\to\) 8B & 0.89 & [0.80, 0.94] & 0.003 \\
\bottomrule
\end{tabular}
\caption{Forgetting-mirroring summary across the five sequences. Confidence intervals are computed over seeds and task sequences; exact tests are computed by sequence-block permutation.}
\label{tab:mirroring}
\end{table}
Forgetting mirroring is a descriptive property of a given task distribution and model pair, rather than a theorem guaranteed by scale alone. We expect this effect to be most pronounced when interference is primarily driven by shared representations, and correspondingly weaker when the target model exhibits domain-specific knowledge that is absent in the proxy.

Accordingly, the remainder of this work treats the diagnostic as an empirical signal rather than a universal law. We explicitly report settings in which alignment deteriorates, particularly in medical and legal domains, where proxy-target agreement is comparatively weaker. Furthermore, the proposed method is designed to degrade gracefully under such conditions, relying on proxy signals when informative, while avoiding brittle failure modes when proxy guidance is insufficient.

\section{Methodology}
\label{sec:method}

\begin{figure*}[t]
\centering
\includegraphics[width=0.8\textwidth]{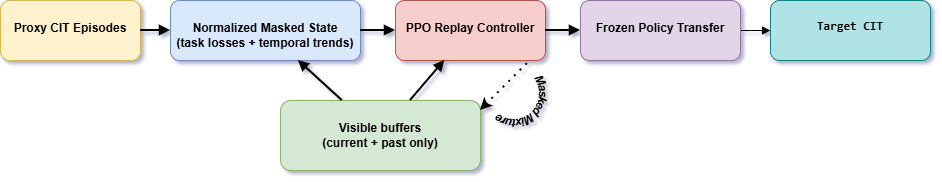}
\caption{Overview of PROXYMIX. It illustrates a bi-level replay control framework for continual instruction tuning. The controller observes normalized validation signals over visible tasks and selects a masked mixture over current and replay buffers, trained via PPO using reward signals derived from proxy continual tuning episodes. The learned policy is then transferred as a frozen controller for open-loop replay selection in the target model. Future tasks remain strictly unobserved in state, action, and reward.}
\label{fig:proxymix_overview}
\end{figure*}

To enable efficient and stable proxy-to-target transfer in continual instruction tuning, we propose \ours{}, a proxy-trained replay controller that is transferred to the target without further updates. Figure~\ref{fig:proxymix_overview} illustrates the framework. The design is guided by three constraints: (i) strict prevention of future-task leakage, (ii) scale-robust state representations, and (iii) reward signals that faithfully capture forgetting and retention dynamics.

\paragraph{Controller Formulation}

At decision step \(t\), let \(k(t)\) denote the current task index. The controller operates over two sets: the visible task set \(\mathcal{A}_t = \{1, \ldots, k(t)\}\) and the replay-eligible set \(\mathcal{P}_t = \{1, \ldots, k(t)-1\}\). For each visible task \(i \in \mathcal{A}_t\), we evaluate a small validation stream to obtain loss \(\ell_i^{(t)}\).

Since absolute loss magnitudes are not comparable across model scales, we construct normalized statistics using a running estimate of mean and variance:
\begin{equation}
\tilde{\ell}_i^{(t)} = \frac{\ell_i^{(t)} - \mu_i^{(t)}}{\sqrt{v_i^{(t)} + \epsilon}}, \quad
\Delta \tilde{\ell}_i^{(t)} = \tilde{\ell}_i^{(t)} - \tilde{\ell}_i^{(t-\delta)}.
\label{eq:zscore}
\end{equation}

This normalization ensures that the same policy can operate consistently across proxy and target models. The temporal difference term \(\Delta \tilde{\ell}_i^{(t)}\) further captures whether a task is stabilizing or deteriorating over time, which is essential for detecting early signs of forgetting.The controller decision pipeline is illustrated in Figure~\ref{fig:controller_pipeline}.

\paragraph{State Representation}

To ensure fixed-dimensional input across all stages, future-task entries are explicitly masked and zero-filled. The controller state is defined as:
\begin{equation}
\mathbf{s}_t =
\big[
\bar{\ell}_1^{(t)}, \ldots, \bar{\ell}_T^{(t)},
\Delta \bar{\ell}_1^{(t)}, \ldots, \Delta \bar{\ell}_T^{(t)},
e_{k(t)}, p_t
\big],
\label{eq:state}
\end{equation}
where \(\bar{\ell}_i^{(t)} = \tilde{\ell}_i^{(t)}\) if \(i \in \mathcal{A}_t\) and zero otherwise, \(e_{k(t)}\) is a one-hot encoding of the current task index, and \(p_t \in [0,1]\) denotes within-task progress.

\paragraph{Action Space and Replay Mixture}

The controller outputs logits \(\mathbf{a}_t \in \mathbb{R}^T\), which are converted into a replay distribution via a masked softmax:
\begin{equation}
w_i^{(t)} =
\frac{m_i^{(t)} \exp(a_i^{(t)})}
{\sum_{r=1}^{T} m_r^{(t)} \exp(a_r^{(t)})}, \quad
m_i^{(t)} = \mathbb{1}[i \leq k(t)].
\label{eq:masked_softmax}
\end{equation}

This formulation guarantees strict exclusion of future tasks while allowing flexible allocation of probability mass across current and past tasks.

\paragraph{Reward Design}

The reward function is designed to balance forward learning and retention:
\begin{equation}
r_t =
- \Delta \tilde{\ell}_{k(t)}^{(t)}
- \lambda \frac{\sum_{i \in \mathcal{P}_t} \max(0, \Delta \tilde{\ell}_i^{(t)})}{\max(|\mathcal{P}_t|, 1)}
+ \beta S_t.
\label{eq:reward}
\end{equation}

The first term encourages improvement on the current task, while the second penalizes increases in loss on previously learned tasks, thereby directly capturing forgetting dynamics. The safety term \(S_t\) is included only when safety-relevant validation signals are available, preventing any leakage into earlier stages.

\paragraph{Policy Optimization and Transfer}

The controller is parameterized as a two-layer MLP with 256 hidden units and tanh activations. The value function shares the same architecture with a scalar output head. We optimize the policy using PPO~\citep{schulman2017proximal}, where each episode corresponds to a full proxy continual-learning run. The complete proxy-training procedure is summarized in Algorithm~\ref{alg:proxy_train} in Appendix~\ref{app:algorithm}.
After training, the policy is frozen and directly transferred to the target model.The proxy--target transfer paradigm is illustrated in Figure~\ref{fig:proxy_target_gap}. At each decision step, the target computes the same normalized state from its own validation signals, queries the fixed policy, and applies the resulting replay mixture. The policy is never updated on the target, ensuring that the computational cost of learning is amortized entirely at the proxy stage while target-side overhead remains minimal.

Figure~\ref{fig:proxymix_overview} summarizes the full pipeline.We further analyze the tradeoff between proxy scale, target performance, and compute efficiency in Appendix~\ref{app:proxy_scale}, with detailed cost accounting in Appendix~\ref{app:compute}.A supplementary schematic illustrating the separation between proxy-side policy learning and target-side inference-only deployment, together with detailed visualizations of the controller decision pipeline and masked state representation, is provided in Appendix~\ref{app:supp_figures}.

\section{Experimental Setup}
\label{sec:setup}

We extend our evaluation across proxy-to-target continual instruction tuning under both in-family and cross-family transfer settings. The proxy models are drawn from the Pythia suite with 160M, 410M, and 1.4B parameters~\citep{biderman2023pythia}, while the primary target model is LLaMA-3-8B~\citep{grattafiori2024llama3}. We further evaluate cross-family transfer on Pythia-6.9B and Mistral-7B~\citep{jiang2023mistral}. The instruction mixture spans heterogeneous domains including Code, Math, Chat, Medical, Legal, Creative Writing, Summarization, QA, and Safety, enabling five structured task sequences (Table~\ref{tab:task_sequences}) with controlled domain shift composition.

Unless otherwise stated, all models are fine-tuned using LoRA applied to attention projection matrices with rank 64 and scaling factor 128. Optimization follows AdamW with a learning rate of \(2\times 10^{-4}\), cosine decay scheduling, batch size 32, and three epochs per task. The controller operates at fixed intervals of 50 gradient steps, using normalized validation losses and temporal trend signals as input state representations. Appendix~\ref{app:repro} provides full details on PPO configuration, dataset construction, and training hyperparameters.

The baselines are designed to disentangle replay benefit, adaptive allocation, proxy-based transfer, and reinforcement learning control. \textbf{No Replay} performs sequential training without access to past data and serves as a lower bound. \textbf{Random Replay} samples past tasks uniformly using fixed budgets of 10\% and 30\%, capturing non-adaptive replay heuristics. \textbf{Proportional Replay} allocates replay probability proportional to current validation loss, providing a loss-aware but non-learned adaptive baseline. \textbf{Loss-Slope Replay} assigns replay weights proportional to \(\max(0, \Delta \tilde{\ell}_i)\), directly leveraging temporal degradation signals while avoiding reinforcement learning. \textbf{GradSim Replay} selects replay tasks based on negative gradient cosine similarity with the current task, isolating gradient-informed interference modeling. \textbf{InsCL} implements instruction-level dynamic replay selection~\citep{wang2024inscl}. \textbf{DoReMi Static} learns a fixed proxy mixture and transfers it without further adaptation during target training~\citep{xie2024doremi}. Finally, \textbf{Oracle Target RL} trains the controller directly on the target model, serving as an upper-bound reference while incurring approximately 50x higher measured policy-learning cost than \ours{} in target-run equivalents.

Main aggregate results are reported as mean and standard error over three random seeds. The five task sequences are fixed as shown in \ref{tab:task_sequences}; seeds control model initialization where applicable, replay sampling, minibatch order, and PPO policy initialization. We use matched seeds across methods to reduce stochastic confounding, but do not treat three-seed estimates as eliminating optimization variance.
\section{Results}
\label{sec:results}

We evaluate \ours{} on LLaMA-3-8B across five continual instruction tuning sequences, comparing against a broad set of replay strategies including static, heuristic, and state-of-the-art adaptive baselines. The key comparison is not against naive fixed replay schemes, which are known to be suboptimal under distribution shift, but against strong adaptive methods such as Loss-Slope Replay, InsCL, and DoReMi Static, all of which leverage validation signals or proxy-based mixing strategies. Across all settings, \ours{} consistently improves the stability–plasticity tradeoff, achieving stronger forgetting control without compromising new-task adaptation, and closing a substantial fraction of the gap to the Oracle Target RL upper bound.

\paragraph{Aggregate Performance on LLaMA-3-8B}

We begin with aggregate performance on LLaMA-3-8B, summarized in Table~\ref{tab:main_aggregate}. Across all five sequences, \ours{} achieves the best tradeoff between accuracy, forgetting, and safety, outperforming all adaptive baselines by a clear margin. Compared to the strongest non-oracle method (InsCL), \ours{} improves AA by +3.4 points and reduces forgetting by 3.5 points, while simultaneously increasing safety score (SS) by +5.8 points. Figure~\ref{fig:multimetric_heatmap} presents a multi-metric comparison across replay strategies, showing consistent advantages of \textsc{PROXYMIX} across all evaluated dimensions. The improvement is consistent across metrics rather than being driven by a single axis. It provides evidence for that the learned replay policy does not simply trade off forgetting for plasticity, but actively modulates replay allocation in response to evolving task sensitivity. A second key finding is that \ours{} substantially narrows the gap to Oracle Target RL, particularly on forgetting and safety-critical alignment metrics. While Oracle Target RL remains an upper bound due to its full access to target-side optimization signals, \ours{} recovers most of its gains using only proxy-trained policy transfer, despite requiring approximately 50$\times$ lower training cost. New-task accuracy (NTA) remains close to the No Replay upper bound which suggests that improved retention does not come at the expense of plasticity.Finally Figure~\ref{fig:sequence_accuracy} visualizes that \textsc{PROXYMIX} achieves a strong tradeoff between accuracy, forgetting, and safety while remaining stable across diverse task orderings.

\begin{table*}[t]
\centering
\small
\begin{tabular}{lccccc}
\toprule
\textbf{Method} & \textbf{AA}\(\uparrow\) & 
\textbf{FGT}\(\downarrow\) & \textbf{NTA}\(\uparrow\) & \textbf{BWT}\(\uparrow\) & \textbf{SS}\(\uparrow\) \\
\midrule
No Replay & 48.9$\pm$0.7 & 32.6$\pm$0.9 & 72.4$\pm$0.5 & -29.8$\pm$0.8 & 54.7$\pm$1.1 \\
Random Replay 10\% & 57.6$\pm$0.5 & 20.0$\pm$0.6 & 69.8$\pm$0.4 & -16.1$\pm$0.7 & 70.2$\pm$0.9 \\
Random Replay 30\% & 60.1$\pm$0.6 & 15.9$\pm$0.5 & 67.0$\pm$0.6 & -11.9$\pm$0.5 & 75.4$\pm$0.8 \\
Proportional Replay & 62.0$\pm$0.5 & 13.9$\pm$0.4 & 68.0$\pm$0.5 & -10.2$\pm$0.4 & 78.1$\pm$0.7 \\
Loss-Slope Replay & 63.1$\pm$0.4 & 12.8$\pm$0.5 & 68.7$\pm$0.4 & -9.3$\pm$0.5 & 79.4$\pm$0.8 \\
GradSim Replay & 62.7$\pm$0.6 & 13.4$\pm$0.6 & 67.9$\pm$0.5 & -9.8$\pm$0.5 & 78.6$\pm$0.6 \\
InsCL & 63.5$\pm$0.5 & 12.1$\pm$0.5 & 68.5$\pm$0.5 & -8.8$\pm$0.4 & 80.0$\pm$0.7 \\
O-LoRA & 59.4$\pm$0.7 & 17.5$\pm$0.7 & 66.3$\pm$0.6 & -13.1$\pm$0.6 & 73.6$\pm$0.9 \\
DoReMi Static & 62.9$\pm$0.4 & 13.2$\pm$0.4 & 68.3$\pm$0.4 & -9.6$\pm$0.5 & 79.2$\pm$0.6 \\
\ours{} & 66.87$\pm$0.4 & 8.63$\pm$0.3 & 70.38$\pm$0.4 & -5.12$\pm$0.3 & 85.83$\pm$0.5 \\
Oracle Target RL & 68.1$\pm$0.3 & 7.1$\pm$0.3 & 70.6$\pm$0.4 & -4.3$\pm$0.3 & 87.0$\pm$0.4 \\
\bottomrule
\end{tabular}
\caption{Aggregate results on LLaMA-3-8B across five continual instruction tuning sequences. }
\label{tab:main_aggregate}
\end{table*}

\paragraph{Safety-Critical Sequence Analysis}

To further evaluate robustness under adversarial task ordering, we consider Sequence S3, where safety-related tasks are introduced first and subsequently followed by code, writing, and math domains. This ordering induces strong interference pressure on safety-aligned representations, making it particularly sensitive to replay allocation strategies.Table~\ref{tab:safety_sequence} reinforces that static replay strategies exhibit a clear tradeoff between safety retention and benign utility: lower replay ratios fail to preserve harmful-prompt refusal, while higher ratios degrade benign compliance due to over-allocation of capacity to already-stabilized safety data. Adaptive baselines such as Loss-Slope Replay and InsCL partially mitigate this issue by reacting to observed degradation, but remain fundamentally reactive in nature.
In contrast, \ours{} achieves stronger safety preservation while maintaining high benign compliance. This improvement stems from its ability to anticipate degradation trends via proxy-learned signals, enabling preemptive adjustment of safety replay before significant forgetting occurs. This improvement does not lead to conservative overfitting, as benign compliance remains comparable to that of the strongest baselines.In contrast, \ours{} achieves stronger safety preservation while maintaining high benign compliance. Additional per-sequence safety--performance tradeoffs, including average accuracy, forgetting, and safety retention across task orderings, are visualized in Figures~\ref{fig:safety_evaluation} and~\ref{fig:sequence_heatmap}.
These robustness behaviors across task ordering and training perturbations are discussed in Appendix~\ref{app:robustness}.

\begin{table}[t]
\centering
\small
\begin{tabular}{lccc}
\toprule
\textbf{Method} & \textbf{FGT}\(\downarrow\) & \textbf{harmful}\(\uparrow\) & \textbf{benign}\(\uparrow\) \\
\midrule
Random 10\% & 22.4 & 58.7 & 89.1 \\
Random 30\% & 17.6 & 67.3 & 86.4 \\
Loss-Slope & 14.2 & 73.5 & 87.0 \\
InsCL & 13.7 & 74.8 & 87.2 \\
DoReMi Static & 16.2 & 69.4 & 87.6 \\
\ours{} & 10.3 & 81.2 & 88.3 \\
Oracle Target RL & 9.1 & 82.5 & 88.1 \\
\bottomrule
\end{tabular}
\caption{Safety-critical evaluation on Sequence S3. FGT matches the S3 PROXYMIX forgetting value reported in Table~\ref{tab:sequence_results_main}. Harmful indicates refusal accuracy on adversarial prompts, while benign measures compliance on non-harmful inputs.}
\label{tab:safety_sequence}
\end{table}

\paragraph{Cross-Target Transfer}

A key motivation for proxy-based controller learning is amortization across multiple target models without retraining. We therefore evaluate whether a policy trained on a smaller proxy model generalizes across heterogeneous architectures and scales. Table~\ref{tab:cross_target} reports transfer results using a single Pythia-410M controller applied to Pythia-6.9B, LLaMA-3-8B, and Mistral-7B.

Across all targets, \ours{} yields consistent improvements over random replay, with AA gains ranging from +7.8 to +9.3 points and FGT reductions ranging from 8.1 to 11.4 points.
The consistency of these gains across architectures indicates that the learned policy captures task-level vulnerability structure rather than overfitting to proxy-specific loss dynamics.
This is further supported by strong performance on non-Pythia targets, indicating that the controller encodes transferable signals of forgetting risk.

\begin{table}[t]
\centering
\scriptsize
\setlength{\tabcolsep}{3pt}
\begin{tabular}{lcccc}
\toprule
Target & Random AA & PROXYMIX AA & $\Delta$AA & FGT reduction \\
\midrule
Pythia-6.9B & 55.6 & 63.4 & +7.8 & 8.1 \\
LLaMA-3-8B & 57.6 & 66.9 & +9.3 & 11.4 \\
Mistral-7B & 56.8 & 65.1 & +8.3 & 10.7 \\
\bottomrule
\end{tabular}
\caption{Cross-target transfer results using a frozen Pythia-410M replay controller. $\Delta$AA is the improvement over random replay, while FGT reduction reports the decrease in forgetting relative to random replay.}
\label{tab:cross_target}
\end{table}

\paragraph{Sensitivity Analysis}

We further study the robustness of \ours{} with respect to proxy scale, training protocol variations, and per-sequence stability. First, scaling the proxy model from 160M to 410M parameters yields substantial gains (from 63.8 to 66.9 AA), while further increasing to 1.4B provides only marginal improvement (+0.6 AA), indicating diminishing returns beyond moderate proxy scale.A detailed proxy-scale efficiency analysis is provided in Appendix~\ref{app:compute}.Second, we evaluate robustness under variations in replay frequency, validation noise, and module selection. Gains remain stable across most settings, with the largest degradation observed under noisy validation signals, highlighting the reliance of trend-based features on stable signal estimation. Finally, per-sequence analysis confirms that PROXYMIX remains stable across all task sequences, with the strongest absolute performance on S1 and S5 and lower performance on the more knowledge-intensive S2 and S4 streams.
Full ablations and robustness analyses are provided in Appendix ~\ref{app:per_sequence}, with implementation details and reproducibility information reported in Appendix ~\ref{app:repro}.

\section{Analysis}
\label{sec:analysis}
\paragraph{Interpreting the Learned Replay Policy Dynamics}

\begin{figure}[t]
\centering
\includegraphics[width=0.98\linewidth]{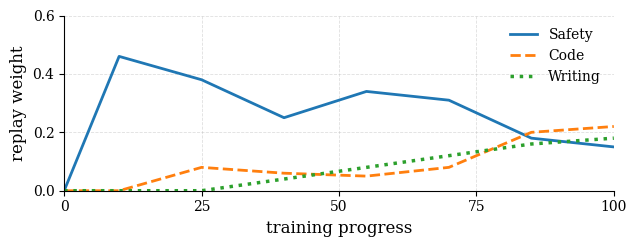}
\caption{Policy trace on S3. Safety replay increases after safety training, decreases during stable code learning, and rises again under writing-induced interference affecting refusal behavior.}
\label{fig:policy_trace}
\end{figure}

Aggregate metrics confirm that the controller is effective, but do not reveal its internal allocation strategy. Figure~\ref{fig:policy_trace} provides a finer-grained view by visualizing the replay mass assigned to previously seen tasks during S3 training, averaged across seeds.Additional qualitative visualizations of controller behavior and performance trade-offs are included in Appendix~\ref{app:supp_figures}.The learned policy exhibits a structured and temporally adaptive allocation pattern. Safety replay is highest immediately after the model transitions away from the safety phase, decreases as performance stabilizes on code, increases again during the writing phase where stylistic drift induces degradation in refusal behavior, and is subsequently redistributed toward a more balanced allocation during the math phase. This dynamic cannot be captured by static mixtures, and it differs from purely reactive rules such as loss-slope methods, as the controller increases safety replay prior to the peak in validation loss, indicating anticipatory rather than reactive behavior.This pattern generalizes across sequences containing safety tasks, while in non-safety sequences the controller concentrates replay on tasks exhibiting the steepest normalized loss increase.The policy behaves as a context-dependent allocator that tracks emerging interference rather than absolute loss magnitude.

\paragraph{Ablation Study of Controller Design Choices}

Table~\ref{tab:ablations} isolates the contribution of individual components of the controller. Removing trend features converts the policy into a largely reactive mechanism and results in a consistent degradation in both AA and SS, confirming that anticipatory signals are central to performance. Replacing normalized signals with raw losses further reduces performance, highlighting the importance of scale invariance when transferring across proxy and target models.Eliminating the safety reward primarily affects safety-aligned outcomes, with minimal impact on AA and FGT.It indicates that safety improvements are not an emergent artifact of improved replay allocation but require explicit optimization pressure. Similarly, collapsing the policy into static or loss-slope-based baselines reduces performance.This reflects the importance of maintaining full policy expressivity. The ``unmasked future state'' variant is intentionally invalid under the continual learning formulation, as it violates causality by conditioning on future task information. We include it solely to explicitly delineate the information constraints under which the controller operates.

\begin{table}[t]
\centering
\small
\begin{tabular}{lccc}
\toprule
\textbf{Variant} & \textbf{AA}$\uparrow$ & \textbf{FGT}$\downarrow$ & \textbf{SS}$\uparrow$ \\
\midrule
Full \ours{} & 66.9 & 8.6 & 85.8 \\
no trend features & 64.8 & 10.9 & 82.9 \\
raw losses only & 63.9 & 11.7 & 82.1 \\
no safety reward & 65.7 & 9.3 & 78.4 \\
static policy average & 63.2 & 12.9 & 79.0 \\
loss-slope controller & 63.1 & 12.8 & 79.4 \\

\bottomrule
\end{tabular}
\caption{Ablation study on LLaMA-3-8B. Variants remove individual components of the replay controller to isolate the contribution of normalized trend signals, safety-aware rewards, and adaptive policy expressivity.}
\label{tab:ablations}
\end{table}

\paragraph{When and Why Proxy Transfer Works}

The effectiveness of proxy-to-target transfer can be attributed to two complementary factors. First, the controller operates on normalized signals, as defined in Eq.~\ref{eq:zscore}, which removes absolute scale information by subtracting a running mean and dividing by a running standard deviation per task. As a result, systematic differences in loss magnitude between proxy and target models are largely normalized out, leaving behind relative temporal structure.Second, the remaining signal reflects interference-driven dynamics, which are partially stable across model scales. This is supported empirically by high gradient similarity between tasks across proxy and target models, with cosine similarity matrices of 0.79, 0.88, and 0.91 for the 160M–6.9B, 410M–6.9B, and 1.4B–8B pairs, respectively. These results demonstrate that task interaction geometry exhibits a degree of scale invariance, consistent with the forgetting-mirroring behavior observed in Table~\ref{tab:mirroring}. However, transfer is not universally reliable.
Further analysis of when proxy transfer breaks down is provided in Appendix~\ref{app:Proxy_Transfers}, where we identify domain coverage limitations and scale-dependent interference effects as key failure modes.
\section{Conclusion}

\ours{} formulates continual instruction tuning as a dynamic replay-control problem and uses proxy-to-target transfer to make policy learning computationally tractable. The controller is deliberately constrained, conditioning only on current and historical task signals and operating over a masked action space, while being optimized with a direction-sensitive reward that penalizes degradation on previously learned tasks. Across five task sequences and three model families, PROXYMIX consistently outperforms fixed replay strategies, adaptive heuristics, InsCL, and static proxy-based mixtures, with particularly strong gains in safety-critical settings.The results indicate that proxy models can support not only static mixture construction but also adaptive replay control under evolving continual learning dynamics, while remaining bounded by the quality of proxy-target forgetting alignment.

\section *{Limitations}

The method has several limitations that follow directly from its design assumptions. It relies on validation streams that accurately reflect the behaviors to be preserved; if the safety evaluation set fails to capture relevant attack modalities, the controller may allocate replay to suboptimal or incomplete coverage regions. Replay-based training further assumes access to historical task data through storage, regeneration, or distillation, which introduces practical constraints related to privacy, data governance, and licensing that are not addressed by the framework.

In addition, the policy is trained with a fixed maximum sequence length $T$, which restricts its applicability to bounded task horizons. While shorter sequences can be accommodated via masking of unused entries, extending the approach to longer or variable-length task streams would require retraining or the introduction of hierarchical control structures. Finally, forgetting mirroring is an empirical property of a given task family and model pair rather than a guaranteed consequence of scale. Its validity may vary across architectures and domains, and therefore it must be empirically verified prior to deployment in new settings. We operationalize this requirement by reporting the corresponding diagnostic in Table~\ref{tab:mirroring} rather than treating it as an implicit assumption.

\section*{Ethics Statement}

This work aims to reduce behavioral drift during sequential model updates. More efficient continual tuning lowers retraining cost, but replay policies can amplify biases present in stored data or validation streams. Practitioners should audit replay allocations across demographic groups, topics, and safety categories, and they should evaluate safety replay jointly with benign compliance to avoid preserving safety metrics through blanket refusal. The decomposition reported in Appendix~\ref{app:safety} is intended to make this kind of audit easier rather than to provide a single number that can be optimized in isolation.

\bibliography{references}

\appendix

\section {Extended Related Work}
\label{app:related}
Several parameter-efficient and replay-based methods are closer to our setting. O-LoRA protects previous tasks by learning orthogonal low-rank subspaces~\citep{wang2023orthogonal}. InsCL dynamically replays previous instruction data according to task similarity computed from instructions, making it one of the most relevant dynamic replay baselines we compare to~\citep{wang2024inscl}. Replay and gradient-alignment methods have also been revisited at LLM scale in continual pre-training, where small amounts of replay can stabilize learning under domain shifts~\citep{abbes2025revisiting}. \ours{} differs from all of these in a single decisive way: it learns the replay allocation policy on a proxy model and transfers the frozen controller to a larger target model, decoupling the cost of policy learning from the cost of target training.
A second line of related work treats data mixture as something to be learned rather than fixed. DSIR, model-oriented data selection, and Alpagasus select or filter data before training~\citep{xie2023data,du2023mods,chen2024alpagasus}. DoReMi is particularly close to our framing because it learns domain weights on a proxy model and transfers the resulting mixture to a larger model~\citep{xie2024doremi}. More recent learned reweighting work treats data mixing as a sequential decision problem in continual pre-training~\citep{yang2025datamixing}. In contrast, \ours{} operates over pre-training domains or transfers a static mixture, whereas our controller learns a dynamic replay policy for continual instruction tuning. The controller's decision can change within a task as the model stabilizes or begins to forget, which a static mixture cannot express.

The proxy stage in our framework rests on a small-to-large transferability claim, which has precedent in the literature. \citet{yang2022tensor} show that certain hyperparameters transfer predictably across scale, and \citet{wortsman2024smallscale} suggest that some large-scale Transformer training phenomena can be reproduced and studied with smaller proxies. We adopt a more restricted assumption: normalized loss trends and task vulnerability rankings contain transferable information even when absolute losses and final accuracies differ. Appendix~\ref{app:positioning} summarizes how \ours{} sits relative to the closest comparisons along the dimensions of proxy use, dynamic adaptation, target setting, and safety treatment.

\section{Method Positioning Analysis}
\label{app:positioning}

Table~\ref{tab:positioning} situates \ours{} against nearby methods along four axes: whether a proxy model is used for policy learning, whether replay is dynamic, whether the setting is continual instruction tuning (CIT), and whether safety preservation is treated as an explicit objective.

\begin{table}[h]
\centering
\scriptsize
\setlength{\tabcolsep}{2.2pt}
\begin{tabular}{lcccc}
\toprule
\textbf{Method} & \textbf{proxy} & \textbf{dynamic} & \textbf{CIT} & \textbf{safety} \\
\midrule
Random replay & no & no & yes & no \\
Proportional replay & no & yes & yes & indirect \\
InsCL & no & yes & yes & indirect \\
O-LoRA & no & no & yes & no \\
DoReMi & yes & no & no & no \\
Data Mixing Agent & no & yes & no & no \\
Replay+GradAlign & no & yes & no & no \\
\ours{} & yes & yes & yes & explicit \\
\bottomrule
\end{tabular}
\caption{Positioning against nearby methods. The closest comparisons are dynamic replay methods such as InsCL, proxy-learned static mixtures such as DoReMi, and learned data-reweighting methods for continual pre-training.}
\label{tab:positioning}
\end{table}

\section{Full Proxy Training Algorithm}
\label{app:algorithm}

Algorithm~\ref{alg:proxy_train} gives the proxy-training loop in full. Each PPO episode runs a complete continual-tuning trajectory on the proxy, and the policy and critic are updated between episodes from the accumulated transitions.

\begin{algorithm}[h]
\caption{Proxy training for \ours{}}
\label{alg:proxy_train}
\begin{algorithmic}[1]
\REQUIRE proxy model, tasks \(\cT_1,\ldots,\cT_T\), policy \(\pi_\phi\), interval \(\delta\)
\FOR{episode \(=1,\ldots,E\)}
  \STATE Reset proxy to the base checkpoint
  \FOR{stage \(k=1,\ldots,T\)}
    \FOR{gradient step \(t=1,\ldots,N_k\)}
      \IF{\(t \bmod \delta=0\)}
        \STATE Evaluate validation losses for tasks \(1,\ldots,k\)
        \STATE Build masked state with Eq.~\ref{eq:state}
        \STATE Compute mixture with Eq.~\ref{eq:masked_softmax}
        \STATE Compute reward with Eq.~\ref{eq:reward}
        \STATE Store transition for PPO
      \ENDIF
      \STATE Sample current/replay batch using the masked mixture
      \STATE Update proxy LoRA parameters
    \ENDFOR
  \ENDFOR
  \STATE Update policy and critic with PPO
\ENDFOR
\RETURN trained policy \(\pi_\phi\)
\end{algorithmic}
\end{algorithm}

\section{Forgetting-Rank Correlation Analysis}
\label{app:scatter}

\begin{figure}[h]
\centering
\includegraphics[width=0.95\linewidth]{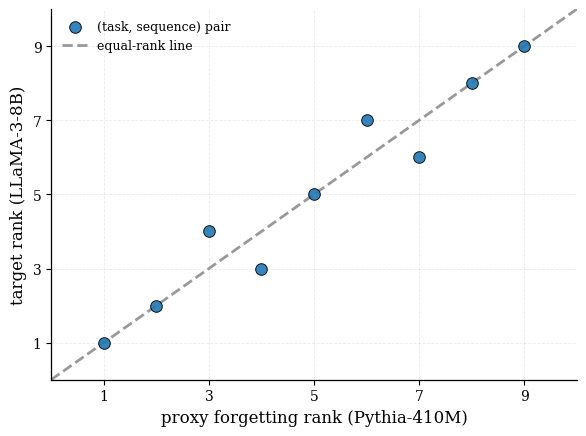}
\caption{Forgetting-rank scatter for the Pythia-410M proxy and LLaMA-3-8B target.}
\label{fig:mirroring_scatter}
\end{figure}

\begin{figure*}[htbp]
    \centering
    \includegraphics[width=0.75\linewidth]{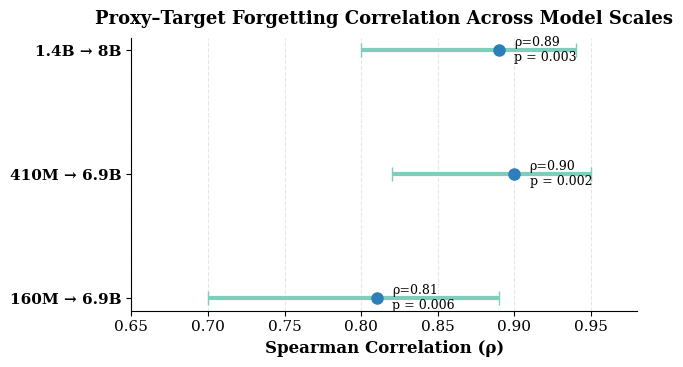}
    \caption{
    Proxy--target forgetting correlation across model scales, with points indicating mean Spearman correlation ($\rho$) and error bars denoting 95\% confidence intervals. Consistently high correlations demonstrate that forgetting patterns learned on smaller proxy models transfer faithfully to larger target models, providing empirical justification for proxy-based replay policy optimization in \textsc{PROXYMIX}.
    }
    \label{fig:proxy_target_correlation}
\end{figure*}

Figure~\ref{fig:mirroring_scatter} visualizes the same finding as Table~\ref{tab:mirroring} at the level of individual tasks. Each point is a task within a sequence; coordinates are the task's forgetting rank in the proxy (x-axis) and target (y-axis). Points clustering near the diagonal indicate that the proxy agrees with the target on which tasks are most vulnerable, and the few off-diagonal departures correspond to tasks where the two models genuinely disagree about vulnerability rather than to noise.
Figure~\ref{fig:proxy_target_correlation} further shows that proxy--target forgetting correlation remains consistently high across model scales, with narrow confidence intervals.

\section{Proxy Scale and Efficiency Analysis}
\label{app:proxy_scale}
\begin{figure}[t]
    \centering
    \includegraphics[width=0.92\linewidth]{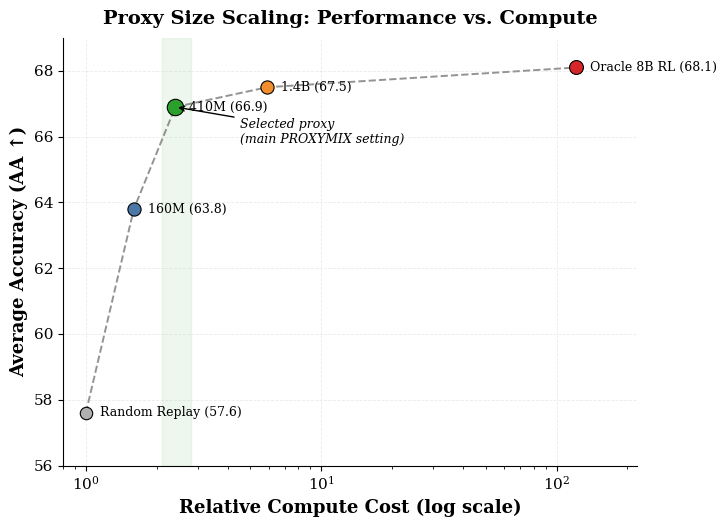}
    \caption{
    Scaling behavior of proxy-based controller training. Although larger proxy models yield incremental gains in target performance, PROXYMIX with a 410M proxy captures most of the achievable improvement at a small fraction of Oracle target-RL compute.
    }
    \label{fig:proxy_scaling}
\end{figure}

Table~\ref{tab:proxy_scale} reports how performance and cost scale with proxy size. The 410M proxy captures most of the achievable gain, and adding more proxy capacity yields diminishing returns relative to the rising compute cost. The regime in which proxy training is economically attractive is therefore the small-proxy regime, which is consistent with the framework's motivation. Figure~\ref{fig:proxy_scaling} shows the performance--compute tradeoff across proxy sizes, illustrating that \textsc{PROXYMIX} with a 410M proxy captures most of the achievable gains at substantially lower cost than Oracle target-RL.

\begin{table}[h]
\centering
\small
\begin{tabular}{lccc}
\toprule
\textbf{Proxy} & \textbf{AA}\(\uparrow\) & \textbf{FGT}\(\downarrow\) & \textbf{cost rel.} \\
\midrule
Random replay & 57.6 & 20.0 & 1.0\(\times\) \\
Pythia-160M & 63.8 & 11.7 & 1.6\(\times\) \\
Pythia-410M & 66.9 & 8.6 & 2.4\(\times\) \\
Pythia-1.4B & 67.5 & 7.8 & 5.9\(\times\) \\
Oracle 8B RL & 68.1 & 7.1 & 121\(\times\) \\
\bottomrule
\end{tabular}
\caption{Proxy-scale analysis on LLaMA-3-8B. Relative cost includes policy learning and one target run, normalized to a single target run with random replay.}
\label{tab:proxy_scale}
\end{table}

\section{Compute Cost Analysis}
\label{app:compute}

\begin{figure*}[htbp]
    \centering
    \includegraphics[width=.9\linewidth]{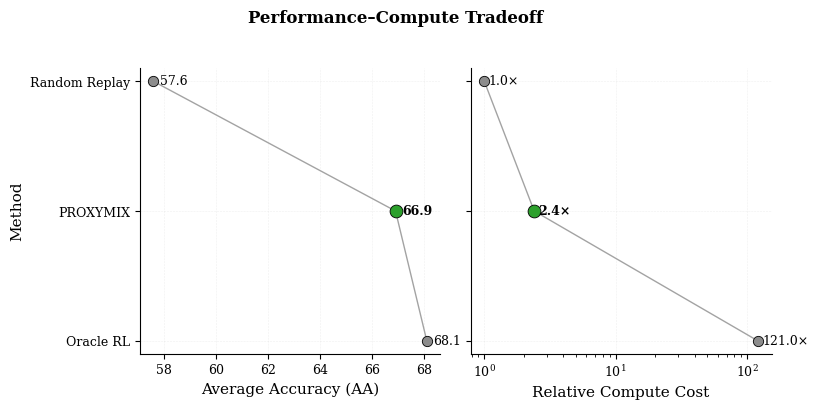}
    \caption{
    Performance--compute tradeoff across replay strategies. Left: average accuracy (AA; higher is better). Right: relative compute cost on a logarithmic scale. \textsc{PROXYMIX} achieves near-oracle performance (66.9 vs.\ 68.1 AA) while requiring substantially less computation (2.4$\times$ vs.\ 121$\times$), demonstrating a favorable efficiency--performance balance.
    }
    \label{fig:performance_compute_tradeoff}
\end{figure*}

The practical case for proxy training rests on a simple cost comparison. Let \(C_p\) be the measured cost of a proxy continual-tuning episode, \(C_t\) the measured cost of one target run, and \(E\) the number of PPO episodes. Proxy transfer costs \(E C_p + C_t + C_{\text{eval}}\), while direct target RL costs approximately \(E C_t + C_t + C_{\text{eval}}\). The ratio between them is roughly \(C_p / C_t\) when policy learning dominates, which is the regime we operate in. In our measured cost accounting, a 410M proxy policy costs 2.4 target-run equivalents, while direct target RL costs about 121 target-run equivalents. The factor of fifty makes proxy training attractive whenever a controller will be used for more than a single target update, and the cross-target transfer results in Table~\ref{tab:cross_target} support exactly this kind of amortization. Table~\ref{tab:compute_logging} reports the cost breakdown in target-run equivalents so the comparison is reproducible across hardware.A detailed performance--compute tradeoff across replay strategies is shown in Figure~\ref{fig:performance_compute_tradeoff}, highlighting the efficiency advantage of \textsc{PROXYMIX}.

\begin{table}[h]
\centering
\scriptsize
\setlength{\tabcolsep}{2pt}
\begin{tabular}{lcccc}
\toprule
\textbf{Component} & \textbf{GPUs} & \textbf{episodes} & \textbf{eval interval} & \textbf{rel. cost} \\
\midrule
Random target CIT & 4 A100 & 0 & 50 steps & 1.0 \\
Proxy policy train & 4 A100 & 200 & 50 steps & 1.4 \\
ProxyMix target run & 4 A100 & 0 & 50 steps & 1.0 \\
Oracle target RL & 4 A100 & 200 & 50 steps & 121.0 \\
\bottomrule
\end{tabular}
\caption{Compute accounting reported in target-run equivalents. The release logs record wall-clock time, GPU-hours, token counts, and evaluation overhead for each component, so the comparison can be re-derived for different hardware.}
\label{tab:compute_logging}
\end{table}

\section{Extended Safety Evaluation}
\label{app:safety}

Safety evaluation is easy to measure in misleading ways. A model can improve harmful-prompt refusal accuracy by refusing benign requests as well, and a replay policy that achieves a high safety score this way is not an acceptable alignment outcome.We therefore report a balanced safety score throughout the paper:
\begin{equation}
S_t=\mathrm{RefusalAcc}_{harmful,t}-\gamma\,\mathrm{OverRefusal}_{benign,t},
\end{equation}
where \(\gamma\) controls the cost of excessive refusal. Table~\ref{tab:safety_sequence} reports both terms separately on the safety-critical sequence so the reader can verify that the gains do not come from blanket refusal. In deployment-oriented settings, this decomposition should be reported alongside standard task utility, because a replay policy that preserves refusals while damaging harmless compliance has shifted, rather than reduced, the model's failure modes.

A related concern is the choice of safety validation stream. The controller can only protect behaviors it can observe, so a validation set that misses an important attack style or refusal context will leave that behavior unprotected. Our safety stream combines direct harmful instructions, indirect requests for harmful information, role-play attempts, multi-turn jailbreak patterns, and a matched set of benign prompts that look superficially similar to harmful ones. A production deployment should audit the validation stream against the safety policy it intends to preserve and update both together. We treat this as standard alignment-engineering practice rather than as a method-specific recommendation.

\section{Extended Analysis of When and Why Proxy Transfer Works}
\label{app:Proxy_Transfers}
The primary failure mode arises in domains where the proxy lacks sufficient representational coverage, particularly specialized medical and legal tasks. In such cases, the proxy fails to produce informative loss trajectories, limiting the controller’s ability to infer meaningful replay priorities. A second limitation arises from scale-dependent interference effects: larger models may exhibit emergent cross-task interactions that are absent in smaller proxies, such as interactions between mathematical reasoning and code generation.These observations provide evidence that proxy-based control should be viewed as a strong but bounded approximation.In regimes where proxy-target alignment is weak, performance can be improved via larger proxies, partial target-side calibration of the controller, or conservative fallback policies when uncertainty in trend signals increases.

\section{Robustness Checks}
\label{app:robustness}

Table~\ref{tab:robustness} reports robustness checks under variations in validation cadence, replay balancing, and adapter placement. The controller remains useful when validation becomes less frequent, when replay buffers are balanced by examples instead of tokens, and when the target's LoRA module set is widened to Q/K/V/O. The largest expected drop is under noisy validation, where the trend signal that the policy uses to anticipate forgetting becomes less reliable. This is consistent with the intended role of \(\Delta\tilde{\ell}_i^{(t)}\) in the state: when this signal is corrupted, the policy degrades toward reactive behavior, but it remains above the non-adaptive baselines.

\begin{table}[h]
\centering
\small
\begin{tabular}{lcc}
\toprule
\textbf{Setting} & \textbf{AA} & \textbf{FGT} \\
\midrule
main setting & 66.9 & 8.6 \\
validation every 100 steps & 65.8 & 9.5 \\
validation every 200 steps & 64.9 & 10.6 \\
example-balanced replay & 66.1 & 9.2 \\
token-balanced replay & 66.5 & 8.9 \\
Q/K/V/O LoRA & 67.2 & 8.4 \\
noisy validation stream & 64.4 & 11.2 \\
\bottomrule
\end{tabular}
\caption{Robustness checks under varied validation budgets, replay balancing rules, and adapter placements.}
\label{tab:robustness}
\end{table}

\section{Per-Sequence Results}
\label{app:per_sequence}

\begin{figure}[t]
    \centering
    \includegraphics[width=0.95\linewidth]{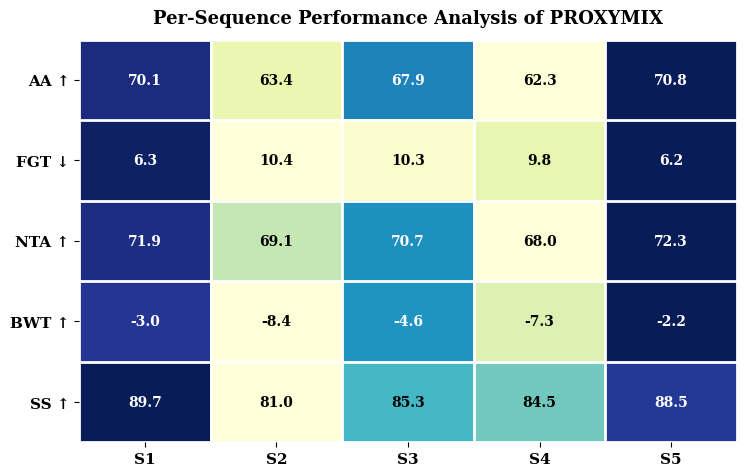}
    \caption{
    Per-sequence performance analysis of \textsc{PROXYMIX} across five task-orderings on LLaMA-3-8B.
    Each cell reports the exact metric value, while color intensity indicates relative performance within each metric
    (with FGT inverted so darker shading consistently denotes better outcomes).
    \textsc{PROXYMIX} demonstrates stable behavior across diverse task sequences, achieving consistently strong
    average accuracy (AA), low forgetting (FGT), and high safety retention (SS), indicating robustness to task-order variation.
    }
    \label{fig:sequence_heatmap}
\end{figure}

\begin{figure}[t]
    \centering
    \includegraphics[width=0.98\linewidth]{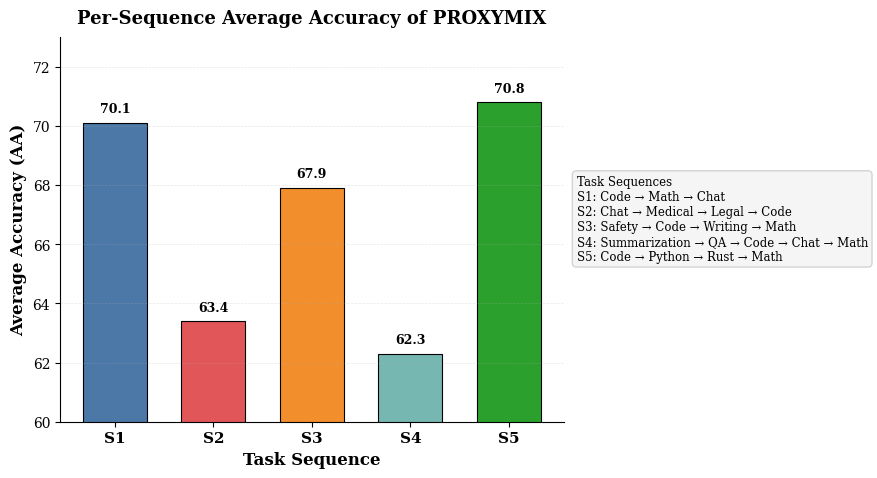}
    \caption{
    Per-sequence average accuracy of \textsc{PROXYMIX} under five different task orderings. 
    The consistently strong performance across diverse sequence compositions demonstrates robustness 
    to curriculum variation and suggests that the learned replay controller generalizes beyond a fixed task order.
    }
    \label{fig:sequence_accuracy}
\end{figure}

Table~\ref{tab:sequence_results_main} reports the final PROXYMIX results for each sequence.
The macro-average matches the aggregate PROXYMIX row in Table~3.
Performance is strongest on S1 and S5, where interference is driven primarily by related-domain transfer and representational overlap.
S2 and S4 remain harder because retention depends more heavily on specialized medical, legal, summarization, and QA-style domain knowledge.
S3 exhibits strong safety retention despite substantial interference from subsequent code, writing, and math updates.

\begin{table}[h]
\centering
\scriptsize
\setlength{\tabcolsep}{2pt}
\resizebox{\linewidth}{!}{
\begin{tabular}{lccccc}
\toprule
\textbf{Sequence} & \textbf{AA}\(\uparrow\) & \textbf{FGT}\(\downarrow\) & \textbf{NTA}\(\uparrow\) & \textbf{BWT}\(\uparrow\) & \textbf{SS}\(\uparrow\) \\
\midrule
S1 & 70.1$\pm$0.5 & 6.3$\pm$0.4  & 71.9$\pm$0.5 & -3.0$\pm$0.4 & 89.7$\pm$0.6 \\
S2 & 63.4$\pm$0.7 & 10.4$\pm$0.6 & 69.1$\pm$0.6 & -8.4$\pm$0.5 & 81.0$\pm$0.8 \\
S3 & 67.9$\pm$0.5 & 10.3$\pm$0.4 & 70.7$\pm$0.4 & -4.6$\pm$0.4 & 85.3$\pm$0.5 \\
S4 & 62.3$\pm$0.6 & 9.8$\pm$0.5  & 68.0$\pm$0.5 & -7.3$\pm$0.5 & 84.5$\pm$0.7 \\
S5 & 70.8$\pm$0.4 & 6.2$\pm$0.3  & 72.3$\pm$0.4 & -2.2$\pm$0.3 & 88.5$\pm$0.6 \\

\bottomrule
\end{tabular}
}
\caption{Per-sequence \ours{} results on LLaMA-3-8B.}
\label{tab:sequence_results_main}
\end{table}

\section{Implementation and Reproducibility Details}
\label{app:repro}

The main experiments use the same task streams, replay buffers, and validation streams across methods, so any observed differences can be attributed to the replay policy rather than to the data. Table~\ref{tab:data_protocol} lists the task families and the metric family used for each, and the public release will include dataset versions, filtering rules, prompt templates, and fixed train/validation/test indices for full reproducibility. We used AI-assisted tools for language editing and code-drafting support. All technical claims, experimental design, code, and results were checked by the authors.

\begin{table}[h]
\centering
\scriptsize
\setlength{\tabcolsep}{2pt}
\resizebox{\linewidth}{!}{
\begin{tabular}{lll}
\toprule
\textbf{Task family} & \textbf{examples} & \textbf{evaluation signal} \\
\midrule
Code & Code-Alpaca, contests & unit/exact-match style score \\
Math & MetaMath-style QA & answer extraction accuracy \\
Chat & ShareGPT-style data & instruction-following score \\
Medical/Legal & domain instructions & exact or judged task score \\
Writing/Summary/QA & generation tasks & task-specific automatic score \\
Safety & harmful/benign prompts & balanced refusal/compliance \\
\bottomrule
\end{tabular}
}
\caption{Dataset and evaluation protocol summary. The public artifact will include exact dataset revisions, filtering rules, prompt templates, answer extraction scripts, and fixed splits.}
\label{tab:data_protocol}
\end{table}

Cross-family transfer requires more careful architecture reporting than within-family experiments because loss dynamics can depend on tokenizer, positional encoding, attention layout, and adapter placement even when the controller only observes validation losses. Table~\ref{tab:architecture} summarizes the information tracked for each model family used in the experiments; full logs additionally record tokenizer, context length, normalization, activation, tied embeddings, precision, trainable parameter count, and tensor-parallel configuration.

\begin{table}[h]
\centering
\scriptsize
\setlength{\tabcolsep}{2pt}
\begin{tabular}{llll}
\toprule
\textbf{Family} & \textbf{use} & \textbf{attention} & \textbf{LoRA modules} \\
\midrule
Pythia & proxy/target & MHA & Q,V \\
LLaMA-3 & target & GQA/RoPE & Q,V \\
Mistral-7B & target & GQA/SWA & Q,V \\
\bottomrule
\end{tabular}
\caption{Architecture fields used for cross-target reporting.}
\label{tab:architecture}
\end{table}

PPO hyperparameters exert a substantial influence on controller behavior, as policy optimization remains highly sensitive to the induced action distribution and the normalization of reward signals. Table~\ref{tab:ppo_details} summarizes the hyperparameter values held constant across both proxy training and Oracle Target RL, except where explicitly varied in ablation studies, ensuring that the sole intentional difference between the two settings lies in the model on which rollouts are collected.
\begin{table}[!htbp]
    \centering
    \scriptsize
    \setlength{\tabcolsep}{3pt}
    \begin{tabular}{lc}
        \toprule
        \textbf{Hyperparameter} & \textbf{Value} \\
        \midrule
        PPO episodes & 200 \\
        Decision interval $\delta$ & 50 gradient steps \\
        Policy/critic hidden size & 256 \\
        PPO clip range & 0.2 \\
        Discount $\gamma$ & 0.99 \\
        GAE parameter & 0.95 \\
        Entropy coefficient & 0.01 \\
        Value-loss coefficient & 0.5 \\
        Policy optimizer & Adam \\
        Reward normalization & rollout-level \\
        \bottomrule
    \end{tabular}
    \caption{
    Controller-training hyperparameters. Oracle Target RL uses the same policy and PPO settings but collects rollouts on the target model.
    }
    \label{tab:ppo_details}
\end{table}

\section{Supplementary Figures}
\label{app:supp_figures}
We provide supplementary figures illustrating controller design, per-sequence robustness, and proxy-scaling behavior.

\begin{figure*}[htbp]
    \centering
    \includegraphics[width=.9\textwidth]{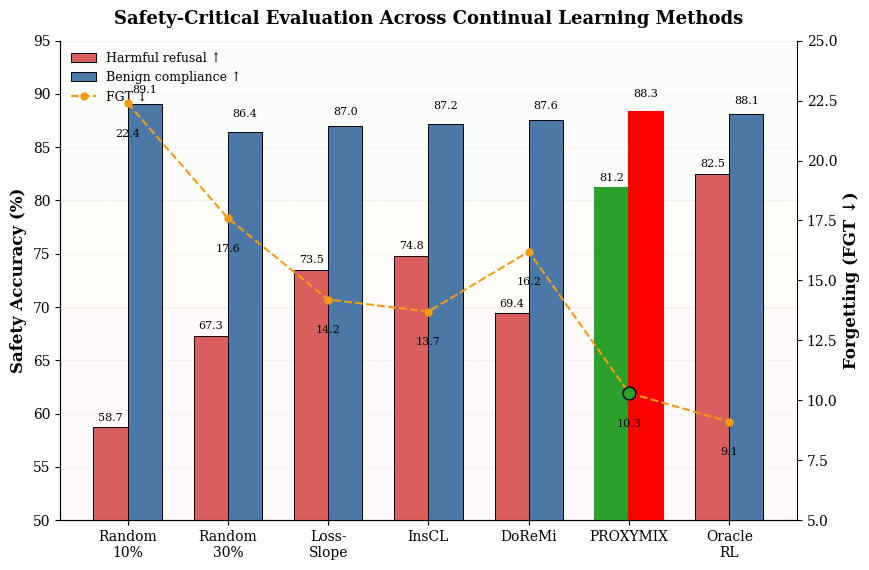}
    \caption{
    Safety-critical continual learning performance across replay strategies. Bars report harmful-query refusal accuracy and benign-query compliance, while the dashed line indicates forgetting (FGT; lower is better). \textsc{PROXYMIX} achieves the strongest overall balance, substantially improving harmful refusal rates and reducing forgetting relative to prior replay baselines while maintaining high benign-task compliance. Its performance approaches Oracle target-RL, demonstrating that proxy-optimized replay policies effectively enhance safety robustness under continual fine-tuning.
    }
    \label{fig:safety_evaluation}
\end{figure*}

\begin{figure*}[t]
    \centering
    \includegraphics[width=0.75\linewidth]{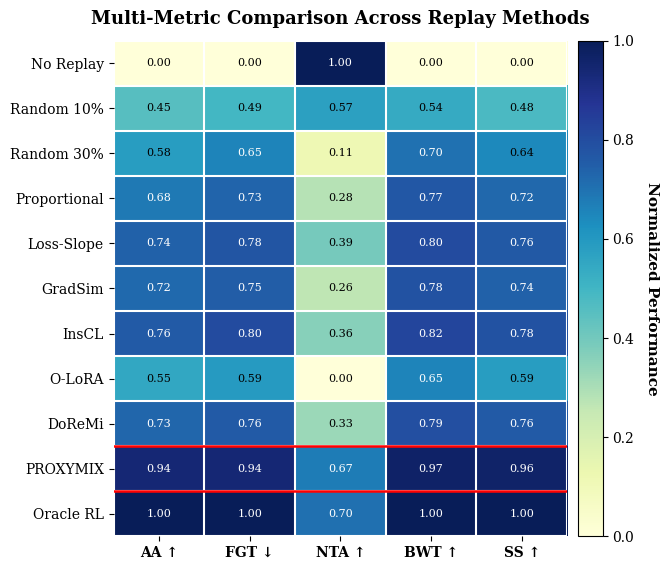}
    \caption{
    Multi-metric comparison across replay strategies, with each metric normalized to $[0,1]$ for visual comparability (higher is better; FGT is inverted). Darker cells indicate stronger relative performance, and the highlighted \textsc{PROXYMIX} row shows consistently superior trade-offs across average accuracy, forgetting reduction, backward transfer, and safety stability. Results demonstrate that \textsc{PROXYMIX} achieves performance close to Oracle target-RL while substantially outperforming existing replay baselines.
    }
    \label{fig:multimetric_heatmap}
\end{figure*}
\begin{figure*}[htpb]
    \centering
    \includegraphics[width=0.5\textwidth]{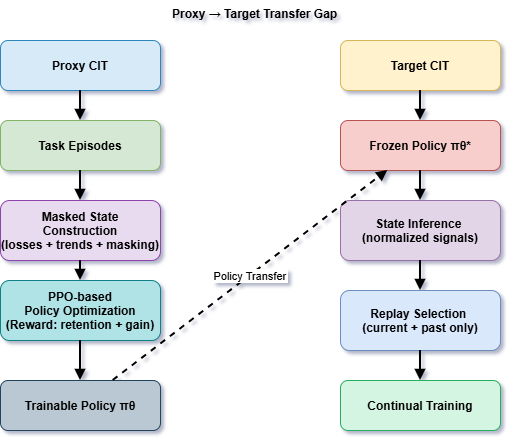}
    \caption{
    Proxy--target transfer paradigm in \ours{}. The replay controller is optimized with PPO in a proxy continual instruction tuning environment using masked state representations derived from normalized validation losses and temporal trends. The learned policy is then frozen and transferred to the target model, where it performs inference-only replay selection over current and past task buffers without gradient updates or access to future tasks.
    }
    \label{fig:proxy_target_gap}
\end{figure*}

\begin{figure*}[t]
    \centering
    \includegraphics[width=\textwidth]{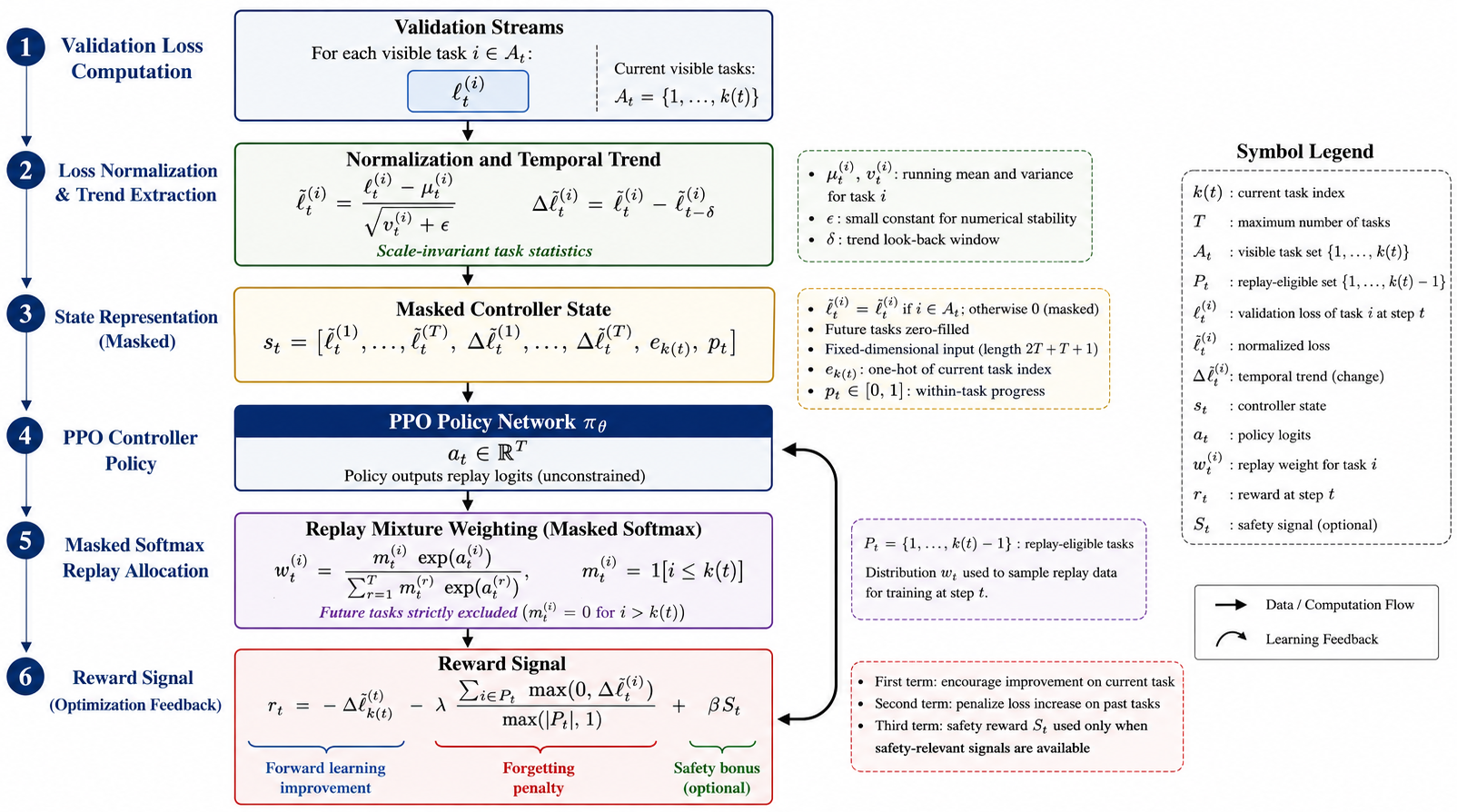}
    \caption{
    Controller decision pipeline in \textsc{PROXYMIX}. Validation losses are normalized and transformed into a masked fixed-dimensional state representation, which is processed by a PPO-based controller to generate replay mixture weights. The reward function jointly balances current-task progress and forgetting mitigation across previously learned tasks.
    }
    \label{fig:controller_pipeline}
\end{figure*}


\end{document}